\documentclass[10pt]{article}
\usepackage[letterpaper]{geometry}
\usepackage{hicss}
\usepackage{times}
\usepackage{hyphenat}
\usepackage{url}
\usepackage{latexsym}
\usepackage{indentfirst}
\usepackage{graphicx}
\graphicspath{{images/}}
\usepackage[
    style=authoryear,
  ]{biblatex}
\usepackage{adjustbox}
\graphicspath{{images/}}
\usepackage{makecell}
\usepackage{wasysym}
\usepackage{multirow}
\usepackage{enumitem}
\setlist[itemize]{noitemsep, topsep=0pt}
\setlist[enumerate]{noitemsep, topsep=0pt}
\setlist{nosep,topsep=-\parskip}

\title{Towards a Benchmark for Large Language Models \\ for Business Process Management Tasks}

 \author{Kiran Busch \\
  Kühne Logistics University \\
  {\underline{kiran.busch@klu.org}} \\ \\ \And
  Henrik Leopold \\
  Kühne Logistics University \\
  {\underline{ henrik.leopold@klu.org} } \\  }

\date{}

\begin{document}
\maketitle
\begin{abstract}

An increasing number of organizations are deploying Large Language Models (LLMs) for a wide range of tasks. Despite their general utility, LLMs are prone to errors, ranging from inaccuracies to hallucinations. To objectively assess the capabilities of existing LLMs, performance benchmarks are conducted. However, these benchmarks often do not translate to more specific real-world tasks. This paper addresses the gap in benchmarking LLM performance in the Business Process Management (BPM) domain. Currently, no BPM-specific benchmarks exist, creating uncertainty about the suitability of different LLMs for BPM tasks. This paper systematically compares LLM performance on four BPM tasks focusing on small open-source models. 
The analysis aims to identify task-specific performance variations, compare the effectiveness of open-source versus commercial models, and assess the impact of model size on BPM task performance. This paper provides insights into the practical applications of LLMs in BPM, guiding organizations in selecting appropriate models for their specific needs.
\end{abstract}


\section{Introduction}
\label{sec:introduction}

An increasing number of organizations are deploying Large Language Models (LLMs), either by developing their own models~\parencite{fedus2022switch, raffel2023exploring} or utilizing readily available ones through APIs, such as OpenAI's GPT-3~\parencite{wolf-etal-2020-transformers}. These LLMs have demonstrated significant utility across a wide range of tasks. They do, however, make mistakes, which can vary from inaccuracies to hallucinations \parencite{shuster2021retrieval, zhou-etal-2021-detecting}. Therefore, to objectively quantify the capabilities of LLMs, so-called performance benchmarks are conducted. Such benchmarks have a long history in Natural Language Processing (NLP) research and typically involve solving a particular task on a dedicated dataset \parencite{marcus1993building, pradhan2012conll}. Common examples include text natural language inference, sentiment analysis, and question answering~\parencite{kiela2021dynabench}.

What these benchmarks have in common is that they focus on rather generic tasks. As a result, LLMs might perform extraordinarily well in the context of benchmarks and might yet fail on simple real-world tasks that are more specific \parencite{kiela2021dynabench}. This raises the question, to what extent the performance results reported from benchmarks can be transferred to more specific domains. One domain that is of particular importance for many organizations is the Business Process Management (BPM) domain. That is, because most organizations employ LLMs in the first place to increase efficiency \parencite{LLMsImpact}. As this, ultimately, impacts the way how business processes are designed and executed, understanding how LLMs perform on typical BPM tasks, such as process analysis or process prediction, is highly relevant~\parencite{grohs2023large}. Currently, however, there are no BPM-specific benchmarks available, leaving it unclear which LLMs are suitable for which BPM task \parencite{busch2023just}. It also remains unclear whether the widely propagated insight that performance generally increases with model size \parencite{brown2020language, kaplan2020scaling} can be transferred to the BPM domain. 

Recognizing this, we use this paper to systematically compare the performance of LLMs on four established BPM tasks. We chose both open-source as well as closed-source LLMs. Among others, this allows us to understand whether there are differences across different tasks, whether open-source and commercial models perform similarly well, and to what extent the size of an LLM affects the performance.   

The rest of the paper is structured as follows. Section~\ref{sec:bpm_tasks} elaborates on the use of LLMs in the context of BPM and introduces the analyzed BPM tasks. Section~\ref{sec:experiments} introduces our experimental setup. Section~\ref{sec:results} discusses the results and Section~\ref{sec:implications} the implications before Section~\ref{sec:conclusion} concludes the paper.



\section{Business process management tasks}
\label{sec:bpm_tasks}

Business Process Management aims to understand, analyze, and improve how work is done in an organization~\parencite{dumas2018fundamentals}. To accomplish this, BPM experts in industry employ a diverse set of means ranging from management methodologies, such as Six Sigma, to specific analytical tools, such as process mining. Since a considerable amount of knowledge pertaining to business processes is captured in data sources that contain text (e.g., work instructions, event logs, e-mails), NLP has become increasingly popular over the last decade in BPM research~\parencite{van2018challenges}. With the introduction of GPT3, the focus has shifted to the use of LLMs~\parencite{busch2023just}. 

To develop an understanding of how LLMs can be effectively used in the BPM domain, we evaluate their BPM-related capabilities. Specifically, we chose four representative tasks, each addressing a specific BPM problem: 

\begin{enumerate}
    \item \textit{Activity recommendation}: This task helps us to assess the LLM's ability to understand and predict logical activity sequences in business processes. 
    \item \textit{Identifying robotic process automation (RPA) candidates}: This task helps us to evaluate the model's capability to differentiate and categorize various types of tasks, highlighting potential automation opportunities.
    \item \textit{Process question answering}: This task tests the LLM's comprehension and information extraction skills from textual process descriptions.
    \item \textit{Mining declarative process models}: This task allows us to examine the model's proficiency in deriving process constraints from natural language.
\end{enumerate}


We briefly explain each BPM task in detail in the following sections.
Note that although we use these BPM tasks in this paper, they are only intended as a starting point for various additional analyses in the BPM area.

\subsection{Activity recommendation}
Activity recommendation is an important task in the context of Business Process Modeling~\parencite{sola2023activity}. It involves predicting a suitable subsequent activity in a process model based on a sequence of already modeled activities. We define each instance in this problem as a tuple $(H, A)$, where $H$ denotes a history of activities already modeled, arranged in a sequence that corresponds to the logical sequence of the process, and $A$ is the next recommended activity to be added to the process model. The objective is to recommend an activity $A$ that suitably follows the sequence $H$.
This task approaches as a sequence prediction problem, where the sequence of past activities ($H$) is used to predict the next activity ($A$). For example, consider the already modeled activity sequence $H=$ [\textit{receive loan application}, \textit{check credit history}, \textit{approve application}]. A suitable next activity could be $A =$ \textit{send approval}.

\subsection{Identifying RPA candidates}
In the context of BPM, the task of identifying RPA candidates involves classifying activities based on the nature of their execution. Each activity is categorized as a \textit{manual task}, a \textit{user task}, or an \textit{automated task}~\parencite{dumas2018fundamentals}. 
A manual task refers to activities where no IT system is involved. A user task involves a human interacting with an IT system, and an automated task is performed entirely by IT systems without human involvement. We define each instance in this problem as a tuple $(E, C)$, where $E$ represents the activity label and $C$ is the classification of the activity into one of the three categories: manual, user, or automated. 
The objective is to accurately classify the activity label $E$ into its respective category $C$. This task is approached as a multi-class classification problem, where the input activity label ($E$) is analyzed to predict its category ($C$). This classification helps in identifying potential RPA candidates by distinguishing between tasks that are currently manual or user-based and those that are automated. For example, given an activity label $E =$ \textit{generate invoice}, the model classifies it as $C =$ \textit{automated task}.

\subsection{Process question answering}
In the context of Business Process Modeling, the task of process question answering involves understanding a textual process description and providing accurate answers to questions related to that description. This task is crucial for assessing the extent to which an LLM comprehends the given process description, which is an essential step for effective process modeling. 
We define each instance in this problem as a triple $(P, Q, A)$, where $P$ represents the textual process description, $Q$ denotes the question related to the process, and $A$ is the expected answer. The objective is to predict the correct answer $A$ given the process description $P$ and the question $Q$. 
This task is approached as a reading comprehension problem, where the input process description ($P$) and question ($Q$) are analyzed to generate the appropriate answer ($A$). This evaluation helps in determining how well an LLM can interpret and extract relevant information from process descriptions. For example, consider the simplified order handling process description $P =$ ``\textit{The process starts with customer order and ends with product delivery.}'' and $Q =$ ``\textit{What is the final step in the process?}''. A correct answer is $A =$ ``\textit{product delivery}''.

\subsection{Mining declarative process models}
Mining declarative process models from textual descriptions is a task that aims to extract flexible and formal process constraints from natural language. Unlike imperative modeling notations, which specify exact sequences of activities, declarative models use constraints to allow more adaptable and knowledge-intensive process definitions. 
In this context, each instance of the problem is defined as a tuple $(T, D)$ where $T$ represents a textual description of a process, and $D$ denotes the set of declarative constraints derived from the text. 
The objective is to accurately generate a set of constraints $D$ that corresponds to the process described in $T$. A way to employ declarative process modeling is the declarative process language \textsc{declare}~\parencite{DiCiccio2022}. We follow the procedure by \cite{grohs2023large} and use the following constraints\footnote{Note that these are only informal examples. For a more comprehensive introduction to \textsc{declare}, we refer to~\cite{DiCiccio2022}.}:

\begin{itemize}
\setlength\itemsep{0em}
\item \textit{Initiation constraint:} $\text{{Init}}(a)$ -- The process starts with activity $a$.
    \item \textit{Termination constraint:} $\text{{End}}(a)$ -- The process ends with activity $a$.
    \item \textit{Precedence constraint:} $\text{{Prec}}({a}, {b})$ -- Activity ${b}$ occurs in the process instance only if preceded by activity ${a}$.
    \item \textit{Succession constraint:} $\text{{Succ}}({a}, {b})$ -- Activity ${a}$ (${b}$) occurs if and only if it is followed (preceded) by activity ${b}$ (${a}$) in the process.
    \item \textit{Response constraint:} $\text{{Resp}}({a}, {b})$ -- If activity ${a}$ occurs in the process, then activity ${b}$ occurs after ${a}$.
\end{itemize}
This task can be approached as a Natural Language Processing problem where rule-based techniques are applied to sentences to identify and generate these constraints.
Consider the textual process description $T =$“\textit{The process begins with registration and concludes with certification issuance.}”. A valid constraint would be $D =$Init(registration).

\section{Experimental Setup}
\label{sec:experiments}
The approach pipeline\footnote{We provide the source code of the implementation under this link: \url{https://github.com/KiriBu10/openLLMinBPM-benchmark}.} for this benchmark involves three steps:
Initially, we collect or create different datasets tailored to each specific BPM task. 
Then we prompt the LLMs with task-specific instructions. This involves carefully creating prompts that align with the requirements of each BPM task to ensure that the LLMs generate relevant and accurate responses. 
Finally, we evaluate the performance of the LLMs. This evaluation is based on predefined metrics that are relevant to the respective BPM tasks. 
The following sections provide a detailed overview of the datasets and LLMs employed, as well as the prompt templates and metrics used.

\subsection{Datasets}
To test the LLMs on the different BPM tasks, we use three task-specific datasets. 

\mypar{Dataset 1}
To evaluate the \textit{activity recommendation} task, we construct a new dataset derived from real-world process models available in the SAP Signavio Academic Models (SAP-SAM) collection~\parencite{sola2022sap}. This collection comprises over one million processes from various domains, represented in different modeling notations and languages\footnote{This number includes duplicates.}.
To ensure the dataset's quality, we implement a series of filtering and cleaning operations. Initially, we exclude vendor-provided examples, as these are likely to be duplicates, in accordance with the guidelines established by \cite{sola2022sap}. We further refine the dataset by selecting only models that utilize the BPMN 2.0 notation and contain English language labels.
Subsequently, we employ several label cleaning techniques to improve label consistency, like removing non-alphanumeric characters, addressing special cases such as line breaks, converting all text to lowercase, and eliminating unnecessary spaces.
From the refined collection, we randomly select 300 process models.
To facilitate the extraction of sequences, we convert the selected process models to event logs. Whenever possible, we extract a sequence of four consecutive activities from each event log. The first three activities in each sequence represent the modeled situation, while the fourth activity represents the optimal subsequent activity. We thus adopt the procedure described by \cite{sola2023activity}. This results in a final set of 288 test samples suitable for evaluating the activity recommendation task.

\mypar{Dataset 2}
For the BPM tasks \textit{RPA candidate identification} and \textit{declarative process model mining}, we use the available datasets from \cite{grohs2023large}, which consist of 424 and 104 test samples, respectively. 

\mypar{Dataset 3} 
For the \textit{process question and answer} task, we create a data set based on four process descriptions: dispatch of goods, recourse, credit scoring, and self-service restaurant from \cite{Camunda2015}. For each process description we create a set of questions and answers that reflect different aspects of the processes. 
We create a total of 15 questions and answers across three levels of complexity: easy, medium and complex. We define the different level heuristically based on the complexity and depth of information required to accurately answer the question.
The easy questions focus on straightforward, factual details, such as ``Who writes the package label for small shipments?''. These questions are designed to assess basic understanding and recall of the process steps. 
Medium questions require a deeper understanding and the ability to sequence events correctly, such as ``\textit{What is the sequence of steps if special shipping is required?}''. These questions test the ability to integrate multiple steps and understand conditional sequences within the processes.
Complex questions involve higher-order thinking and scenario-based problem-solving, such as ``\textit{In what scenarios do you close the case without involving a collection agency?}''. These questions assess the ability to apply knowledge in dynamic contexts and understand exceptions and special conditions. This procedure results in a dataset of 60 samples.

\subsection{Selected language models}
To select suitable LLMs for our benchmark, we make the following heuristic considerations: 
First, we prioritize models that are recently published to ensure that our benchmark incorporates the latest advancements in LLM technology. 
Second, we select lightweight models in terms of parameter size to ensure practicality in real-world applications. Models with fewer parameters are generally more resource-efficient, enabling broader accessibility and easier deployment in various BPM environments.
Third, a significant criterion is the availability of models under open-source licenses. Open-source models allow for transparency, reproducibility, and customization, which are crucial for research and development.
Fourth, to provide a comprehensive evaluation, we include GPT-4 ~\parencite{openai2024gpt4}, a well-known closed-source model. This inclusion allows us to benchmark open-source models against a leading, high-performance model. 
As a result, we select seven LLMs: GPT-4, Phi-3 Medium ~\parencite{abdin2024phi3}, Claude 2 ~\parencite{claude2-alpaca}, Falcon 2 ~\parencite{falcon2_2024}, Mixtral-8x7b ~\parencite{jiang2024mixtral}, Llama 3 ~\parencite{llama3modelcard}, and Yi-1.5 ~\parencite{ai2024yi}.
Table~\ref{tab:models} shows an overview of the selected models. 
\begin{table}[ht]
\centering
    \begin{adjustbox}{width=\columnwidth, center}
    \begin{tabular}{lcccc}
    \hline
    \textbf{LLM}&\textbf{Release date}&\textbf{Params}&\textbf{Context length}&\textbf{Licence}\\ 
    \hline
    \hline
    GPT-4& 2023/03 &1,760B&32k & closed source\\ 
    \hline
    Phi3 medium& 2024/05 & 14B & 128k & MIT\\
    Claude2-13b&2023/10&13B&4k&Meta Llama 2\\
    Falcon2& 2024/05&11B&8192&TII Falcon License 2.0\\
    Mixtral-8x7b&2023/12&46,7B&32k&Apache 2.0\\
    Llama3&2024/04&8B&8192&Meta Llama 3\\
    Yi-1.5&2024/05&9B&4096&Apache 2.0\\
    \hline
    \end{tabular}
    \label{tab:summary}
    \end{adjustbox}
\caption{Selected LLMs.}
\label{tab:models}
\end{table}

\subsection{Prompt templates}
To enable LLMs to perform the BPM task effectively, it is essential to develop task-specific prompt templates that outline how prompts should be structured. This aspect can be particularly challenging, as the performance of LLMs is highly sensitive to the prompt templates used~\parencite{webson2021prompt, perez2021true, zhao2021calibrate}. We identified three widely employed prompt templates as summarized in Table~\ref{tab:prompt-pattern}. 
The \textit{Few-shot example} prompt (i) provides the LLM with a task description followed by three examples. The inclusion of examples serves as a guide to help the model understand the task more concretely and generate responses that align with the demonstrated patterns. 
The \textit{Persona} prompt (ii) involves framing the LLM as a specialist for the forthcoming task. By explicitly defining the model's role, we aim to elicit informed and contextually appropriate answers. 
Finally, the \textit{Step-by-step} prompt (iii) encourages the LLM to approach the task methodically by prompting it to think through the problem step by step. This method is intended to enhance the model's logical reasoning capabilities. Additionally, each prompt template includes an \textit{output instruction} that specifies how the model should format its response, ensuring consistency in the generated outputs.

For the purpose of this paper, we initially focus on the few-shot prompting strategy as this has been shown to yield the best results in various contexts~\parencite{brown2020language, perez2021true}. We, however, also conduct a separate robustness analysis to gain detailed insights into the impact of the prompting strategy.    

\begin{table}[ht]
\centering
    \begin{adjustbox}{width=\columnwidth, center}
    \begin{tabular}{llp{5cm}}
    \hline
    &\textbf{Prompt pattern} & \textbf{Prompt} \\ 
    \hline
    \hline
    i&few-shot example& \{\textit{task description}\} \{\textit{3 examples}\} \{\textit{output instruction}\}\\
    ii&persona& You are a specialist in ... . \{\textit{task description}\} \{\textit{output instruction}\}\\
    iii&step-by-step& \{\textit{task description}\} Let\'s think step by step to solve the problem. \{\textit{output instruction}\}\\
    \hline
    \end{tabular}
    \end{adjustbox}
\caption{Prompt pattern used to prompt the LLMs.}
\label{tab:prompt-pattern}
\end{table}

\subsection{Metrics}
To evaluate the performance of the LLMs on various BPM tasks, we use several metrics.

\mypar{Activity recommendation} 
For this task, we use the average cosine similarity \parencite{li2013distance} between the embeddings of the true next activity and the predicted next activity. This metric assesses how closely the model's recommendation aligns with the expected next activity in the sequence.
Let \( A \) be the true next activity and the model's recommendation be \( \hat{A} \). We embed \( A \) and \( \hat{A} \) using a pre-trained BERT model \parencite{devlin2019bert} to obtain their vector representations \( \mathbf{A} \) and \( \mathbf{\hat{A}} \). 
Based on the cosine similarity \( S_C \) between these two vectors, we calculate the average similarity \( S_{\text{avg}} \) across all samples in the dataset:
\setlength{\abovedisplayskip}{0pt}
\setlength{\belowdisplayskip}{0pt}
\setlength{\abovedisplayshortskip}{0pt}
\setlength{\belowdisplayshortskip}{0pt}
\[
S_{\text{avg}} = \frac{1}{N} \sum_{i=1}^{N} S_C(\mathbf{A}_i, \mathbf{\hat{A}}_i)
\]
where \( N \) is the total number of samples in the dataset.

\mypar{Identifying RPA candidates} 
For this task, we use the standard information retrieval metrics precision (\(\text{prec}\)), recall (\(\text{rec}\)), and F1-score (\(\text{F1}\)) to evaluate the classification accuracy. Each activity is classified into one of three categories: manual, user, or automated. Given the number of true positives (TP), false positives (FP), and false negatives (FN), precision and recall are defined as follows:
\[
\text{prec} = \frac{\text{TP}}{\text{TP} + \text{FP}}, \quad \text{rec} = \frac{\text{TP}}{\text{TP} + \text{FN}}
\]
The F1-Score is defined as the harmonic mean of precision and recall. These three metrics are calculated for each category and also averaged across categories for an overall performance metric.

\mypar{Mining declarative process models} 
For this task, we calculate precision, recall, and F1-score for each type of constraint (precedence, response, succession, initiation, termination) as well as overall metrics across all constraint types.
Given a textual description \( T \) and a set of constraints \( D \), the model predicts a set \( \hat{D} \).
In line with the definitions above, we calculate the metrics for each constraint type \( \tau \): 
\[
\text{prec}_{\tau} = \frac{\text{TP}_{\tau}}{\text{TP}_{\tau} + \text{FP}_{\tau}}, \quad 
\text{rec}_{\tau} = \frac{\text{TP}_{\tau}}{\text{TP}_{\tau} + \text{FN}_{\tau}}
\]
where \( \text{TP}_{\tau} \), \( \text{FP}_{\tau} \), and \( \text{FN}_{\tau} \) are the true positives, false positives, and false negatives for each constraint type. The F1-Score for each type \( \text{F1}_{\tau} \) is again given by the harmonic mean of the respective precision and recall values. 
To quantify the overall performance across all constraint types, we also compute the overall precision, recall, and F1-score based on the total number of true positives, false positives, and false negatives.

\mypar{Process Question Answering} 
To evaluate the performance of the LLMs on the task of process question answering, we use the ROUGE-L~\parencite{lin2004rouge} score, which is a common metric for assessing the quality of generated text in comparison to a reference text. The ROUGE-L score measures the longest common subsequence (LCS) between the predicted answer and the true answer, capturing both precision and recall aspects of text similarity.
Let \( A \) be the true answer and \( \hat{A} \) be the predicted answer. We compute ROUGE-L between \( A \) and \( \hat{A} \) as follows:
\[
\text{ROUGE-L}(A, \hat{A}) = \frac{\text{prec}_{\text{LCS}} \cdot \text{rec}_{\text{LCS}}}{\text{prec}_{\text{LCS}} + \text{rec}_{\text{LCS}}}
\]
where
\[
\text{prec}_{\text{LCS}} = \frac{\text{LCS}(A, \hat{A})}{|\hat{A}|}, 
\text{rec}_{\text{LCS}} = \frac{\text{LCS}(A, \hat{A})}{|A|}
\]
and \( \text{LCS}(A, \hat{A}) \) is the length of the longest common subsequence between \( A \) and \( \hat{A} \), and \( |A| \) and \( |\hat{A}| \) refer to the number of words in \( A \) and \( \hat{A} \), respectively.
To calculate the ROUGE-L score across all samples in the dataset, we compute the score for each sample and then average these scores:
\[
\text{Average ROUGE-L} = \frac{1}{N} \sum_{i=1}^{N} \text{ROUGE-L}(A_i, \hat{A}_i)
\]
where \( N \) is the total number of samples in the dataset, \( A_i \) is the true answer for the \( i \)-th sample, and \( \hat{A}_i \) is the predicted answer for the \( i \)-th sample. This average ROUGE-L score provides a comprehensive measure of how well the LLM can interpret and generate accurate answers based on the process descriptions.

Note that we calculate the metrics with strict adherence to the model's ability to generate output exactly as instructed. Any deviation from the given instructions is considered a failure to fulfill the task. Although this approach has limitations—since the output might be correct but not in the desired format—we consider this procedure essential for benchmarking as it allows us to efficiently evaluate a large number of test cases.

\subsection{Implementation}
We implement our study in Python, using the following models from huggingface ~\parencite{huggingface2020}:
\begin{itemize}
\setlength\itemsep{0em}
    \item \textit{Yi-1.5-9B-Chat-Q4\_K\_M.gguf}
    \item \textit{Phi-3-medium-4k-instruct-Q4\_K\_S.gguf}
    \item \textit{Falcon2-11B.Q4\_0.gguf}
    \item \textit{Claude2-alpaca-13b.Q6\_K.gguf}
    \item \textit{Meta-Llama-3-8B-Instruct.Q4\_K\_M.gguf}
    \item \textit{mixtral-8x7b-instruct-v0.1.Q4\_K\_M.gguf}
\end{itemize}
We use quantization techniques to optimize performance and reduce memory usage. 
We deploy GPT-4 via the API from OpenAI~\parencite{openai2024gpt4}. To ensure reproducibility, we use temperature = 0 for all models. The experiments were conducted using an Nvidia RTX A6000 GPU.

\section{Results}
\label{sec:results}
In this section, we discuss the results of our experiments. We start by providing an overview of the overall results. Then, we take a closer look at each BPM task as well as the inference time and token usage. Finally, we analyze the impact of the employed prompt templates.

\begin{figure}[ht]
\includegraphics[width=\columnwidth, page=1]{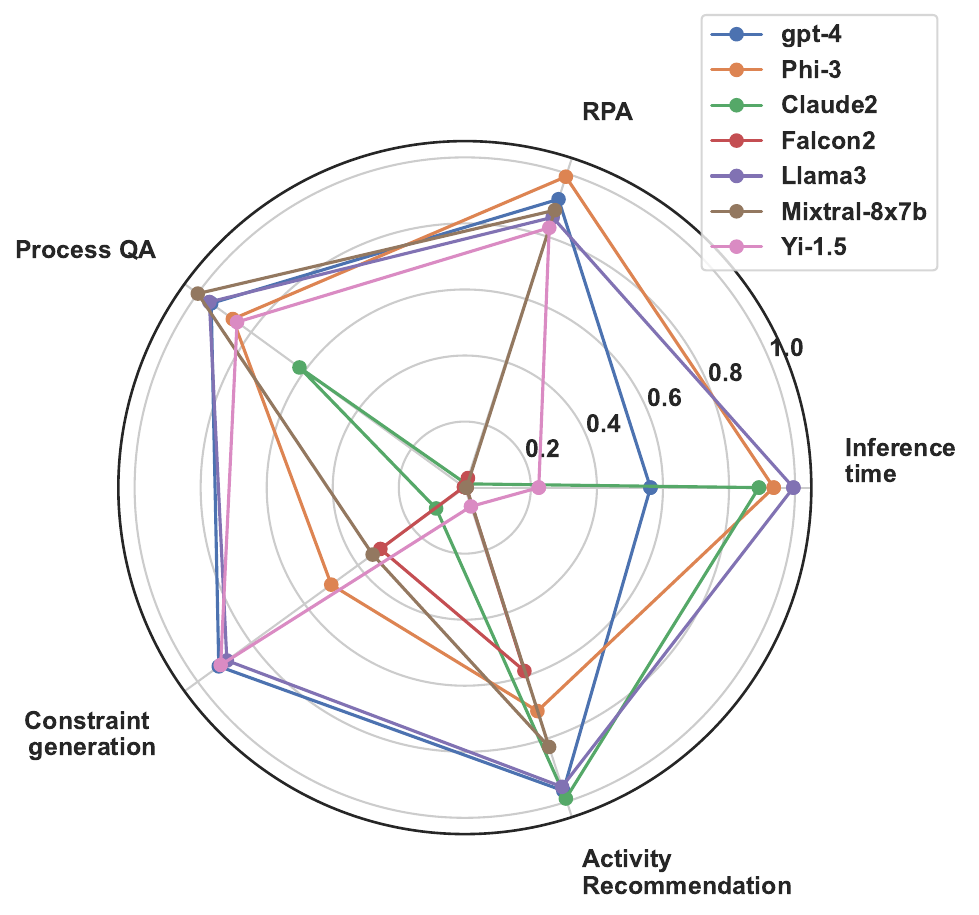}
\caption{Overview of normalized performance results for each BPM task and inference time (with negative sign applied), where higher values indicate better performance. The results are based on few-shot prompting.}
\label{fig:overall-results}
\end{figure}

\mypar{Overall results}
The overall results of our experiments reveal that there are significant performance differences across the four investigated BPM tasks. Figure~\ref{fig:overall-results} provides an overview of the results by showing the normalized performance results for each BPM task as well as the normalized inference time, with a negative sign applied so that higher values represent better performance. In general, we can see that there is no model that performs best in all five dimensions but that the models have different strengths and weaknesses.   


GPT-4 proves to be a robust performer in most tasks and consistently demonstrates high performance, especially in recommending activities and answering process questions. 
Llama3-7b performs very well, particularly in answering process questions and recommending activities, narrowly outperforming GPT-4. 
Phi3-14b and Mixtral-8x7b also perform competitively, particularly in identifying RPA candidates.
Claude2-13b's performance is more specialized. It outperforms in activity recommendation but shows variability across the other tasks.

In summary, all models have unique strengths. However, GPT-4 and Llama3-7b offer the most balanced performance, making them ideal for a wide range of BPM tasks. Phi3-14b and Mixtral-8x7b are strong alternatives for RPA-related tasks, and Claude2-13b outperforms in activity recommendation. Falcon2 and Yi-1.5-9b are more specialized but offer valuable strengths in terms of speed of inference and constraint generation respectively. The results indicate also, that the chosen open-source models can indeed perform comparably to GPT-4 on certain BPM tasks. This finding is noteworthy, given the significant difference in the number of parameters between these models.

In the subsequent sections, we take a more detailed look at the investigated BPM tasks as well as the inference time and token usage.

\mypar{Activity recommendation} The performance results (i.e., the cosine similarity $S_C$) for the activity recommendation task are shown in Table~\ref{tab:activity-recommendation}.
We observe that Claude2-13b achieved the highest average cosine similarity with 0.861, indicating its superior ability to recommend the next activity in the process sequence most accurately among the tested models. GPT-4 and Llama3-7b also performed well, with average cosine similarities of 0.855 and 0.853, respectively.
By contrast, Yi-1.5-9b yielded the lowest performance, with an average cosine similarity of 0.719, highlighting its relatively lower capability in this task compared to the other models.
Overall, the results show that there is a quite a notable performance difference across the tested LLMs. While Claude2-13b exhibited the best performance, models like Phi3-14b and Mixtral-8x7b are, next to GPT-4, only slightly less precise.

\begin{table}[tb]
\small
\centering
\begin{tabular}{lc}
\hline
\textbf{LLM} & \textbf{$\diameter$} $\mathbf{(S_C)}$ \\
\hline
\hline
GPT-4 & 0.855 \\
\hline
Phi3-14b & 0.817 \\
Claude2-13b & \textbf{0.861} \\
Falcon2-11b & 0.798 \\
Llama3-7b & 0.853 \\
Mixtral-8x7b & 0.834 \\
Yi-1.5-9b & 0.719 \\
\hline
\end{tabular}
\caption{Performance results (i.e., the cosine similarity $S_C$) for the activity recommendation task.}
\label{tab:activity-recommendation}
\end{table}

\begin{table*}[bt]
\centering
\begin{adjustbox}{width=\textwidth}
\begin{tabular}{lcccccccccccc}
\hline
 \textbf{LLM} & $\textbf{prec}_\textbf{0}$ & $\textbf{rec}_\textbf{0}$ & $\textbf{F1}_\textbf{0}$  & $\textbf{prec}_\textbf{1}$ & $\textbf{rec}_\textbf{1}$ & $\textbf{F1}_\textbf{1}$ & $\textbf{prec}_\textbf{2}$ & $\textbf{rec}_\textbf{2}$ & $\textbf{F1}_\textbf{2}$ & $\textbf{prec}_\textbf{overall}$ & $\textbf{rec}_\textbf{overall}$ & $\textbf{F1}_\textbf{overall}$ \\
\hline
\hline
GPT-4 & 0.528 & 0.890 & \textbf{0.663} & \textbf{0.886} & 0.580 & 0.701 & 0.174 & 0.250 & \textbf{0.205} & \textbf{0.752} & 0.660 & 0.671 \\
\hline
Phi3-14b & \textbf{0.646} & 0.488 & 0.556 & 0.805& 0.836 & \textbf{0.820} & \textbf{0.250} & 0.062 & 0.100 & 0.736 & \textbf{0.703} & \textbf{0.714} \\
Claude2-13b & 0.462 & 0.236 & 0.312 & 0.389 & 0.025 & 0.047 & 0.035 & \textbf{0.750} & 0.067 & 0.397 & 0.116 & 0.127 \\
Falcon2-11b & 0.301 & \textbf{0.992} & 0.462 & 0.000 & 0.000& 0.000 & 0.000 & 0.000 & 0.000 & 0.090 & 0.297 & 0.138 \\
Llama3-7b & 0.574 & 0.425 & 0.489 & 0.741 & 0.722 & 0.732 & 0.089 & 0.312& 0.139 & 0.666 & 0.618 & 0.636 \\
Mixtral-8x7b & 0.480 & 0.661 & 0.556 & 0.786 & 0.665 & 0.721 & 0.200 & 0.125 & 0.154 & 0.672 & 0.644 & 0.650 \\
Yi-1.5-9b & 0.486 & 0.276 & 0.352 & 0.701 & \textbf{0.858} & 0.771 & 0.000 & 0.000 & 0.000 & 0.610 & 0.651 & 0.616 \\
\hline
\end{tabular}
    \end{adjustbox}
\caption{Performance metrics for identifying RPA candidates across different LLMs.}
  \label{tab:rpa}
\end{table*}

\mypar{Identifying RPA candidates}
Table~\ref{tab:rpa} shows the results for applying the LLMs on the identifying RPA candidates BPM task. It shows precision, recall, and F1-score across three categories: manual (0), user (1), automated (2) tasks as well as the overall performance for each model. We can see that GPT-4 performs robustly across all metrics, achieving a macro-averaged F1-score of 0.671. 
This indicates that GPT-4 has a good balance in identifying various task types.
Phi3-14b achieves the highest macro-averaged F1-score of 0.714, indicating a balanced performance across all categories. This model shows strong precision and recall for user tasks (0.805 and 0.836, respectively), but struggles with automated tasks, as reflected by a low F1-score of 0.100.
Claude2-13b, on the other hand, exhibits a significant discrepancy in its performance. Despite achieving a recall of 0.750 for automated tasks, its overall macro-averaged F1-score is only 0.127, indicating inconsistency in handling other task categories.
Falcon2-11b displays an interesting pattern, with a notably high recall for manual tasks (0.992) but an almost negligible performance in other categories, resulting in a low macro-averaged F1-score of 0.138. This suggests that Falcon2-11b might be overly biased towards detecting manual tasks.
Llama3-7b and Mixtral-8x7b present competitive results, with macro-averaged F1-scores of 0.636 and 0.650, respectively. Both models show moderate strengths and weaknesses across different categories, suggesting a balanced but not exceptional performance.
Finally, Yi-1.5-9b has a strong recall and F1-score for user tasks (0.858 and 0.771, respectively) but fails to identify automated tasks, as indicated by an F1-score of 0. Despite this, it achieves a reasonable macro-averaged F1-score of 0.616.
The findings suggest that, while some LLMs are capable of effectively identifying RPA candidates, there remains a need for further optimization to ensure consistent performance across all task categories. However, the results also again show that open source models can compete with GPT-4 in this task.

\mypar{Process Question Answering}
Table~\ref{tab:process-qa} provides a detailed breakdown of the ROUGE-L scores across different levels of question complexity: easy, medium, and complex.
The overall performance shows that the LLMs can handle questions of varying complexity differently. It is noteworthy that Mixtral-8x7b achieves the highest overall ROUGE-L score of 0.600, outperforming both GPT-4 and Llama3-7b.
For the simple questions, Mixtral-8x7b again showed superior performance with a score of 0.657, closely followed by Llama3-7b and GPT-4, which scored 0.611 and 0.604 respectively. These results indicate that Mixtral-8x7b and Llama3-7b are particularly good at extracting simple, factual details from process descriptions.
For the medium difficult questions, Mixtral-8x7b and Llama3-7b continue to perform good with scores of 0.723 and 0.721 respectively. GPT-4 also performs well with a score of 0.680. This indicates that these models can effectively integrate multiple steps and understand conditional flows within business processes.
For complex questions, which involve higher-order thinking and scenario-based problem-solving, GPT-4 stands out with a score of 0.452. It is followed closely by Mixtral-8x7b and Llama3-7b, which score 0.420 and 0.413, respectively. 
In summary, Mixtral-8x7b consistently performs well across all question complexities, particularly excelling in the easy and medium categories. GPT-4 shows robust performance, especially in handling complex questions, indicating its strong capability in higher-order reasoning within BPM tasks. Llama3-7b also demonstrates strong overall performance, suggesting its effectiveness across various aspects of process comprehension. 
These results emphasis once again that smaller models are quite capable of competing with GPT-4, at least if the questions are not too complex. 
\begin{table*}[tb]
\small
\centering
\begin{tabular}{lrrrr}
\hline
\textbf{LLM} &  $\textbf{ROUGL}_\textbf{easy}$ & $\textbf{ROUGL}_\textbf{medium}$ &$\textbf{ROUGL}_\textbf{complex}$ & $\textbf{ROUGL}_\textbf{overall}$ \\
\hline
\hline
GPT-4 &  0.604 & 0.680 &\textbf{0.452} & 0.579 \\
\hline
Claude2-13b &  0.516 & 0.519 &0.285 & 0.440 \\
Falcon2-11b &  0.201 & 0.163 &0.181 & 0.181 \\
Llama3-7b &  \textbf{0.611} & 0.721 &0.413 & 0.582 \\
Mixtral-8x7b &  0.657 & \textbf{0.723} &0.420 & \textbf{0.600} \\
Phi3-14b &  0.566 & 0.667 &0.401 & 0.545 \\
Yi-1.5-9b &  0.610 & 0.617 &0.387 & 0.538 \\
\hline
\end{tabular}
\caption{Performance metrics for identifying RPA candidates across different LLMs.}
 \label{tab:process-qa}
\end{table*}

\mypar{Mining declarative process models}
Table~\ref{tab:contstraints} shows the performance of the models on mining declarative process models from textual descriptions. We evaluated the models based on overall precision, recall, and F1-score.
The model with the highest precision is Llama3-7b, achieving a precision of 0.375. However, this model struggles with recall, recording only 0.143, which results in an F1-score of 0.207. This indicates that while Llama3-7b is precise in identifying correct constraints, it misses a significant number of them, leading to lower overall effectiveness.
In contrast, Yi-1.5-9b has a recall of 0.619, the highest of all models, and a precision of 0.126, resulting in an F1 score of 0.210, comparable to that of GPT-4 and Llama3-7b. The high recall suggests that Yi-1.5-9b is more effective in identifying a larger number of relevant constraints, although with a lower precision.
GPT-4 shows a precision of 0.135 and a recall of 0.476, leading to the highest F1-score of 0.211. 
Phi3-14b, Claude2-13b, Falcon2-11b, and Mixtral-8x7b show relatively lower precision and recall compared to other models. 
\begin{table}[ht]
\centering
\begin{tabular}{llccc}
\hline
 \textbf{LLM} & \textbf{prec} & \textbf{rec} & \textbf{F1} \\
\hline
\hline
GPT-4 & 0.135 & 0.476 & \textbf{0.211} \\
\hline
Phi3-14b & 0.096 & 0.476 & 0.160 \\
Claude2-13b & 0.068 & 0.333 & 0.113 \\
Falcon2-11b & 0.091 & 0.286 & 0.138 \\
Llama3-7b & \textbf{0.375} & 0.143 & 0.207 \\
Mixtral-8x7b & 0.090 & 0.333 & 0.141 \\
Yi-1.5-9b & 0.126 & \textbf{0.619} & 0.210 \\
\hline
\end{tabular}
\caption{Performance metrics for mining declarative process models from
textual descriptions.}
\label{tab:contstraints}
\end{table}

\mypar{Inference time and token usage}
Both inference time and token usage are important factors to consider for the application of LLMs. While inference time may decide about the overall applicability of a model (as users might only be willing to wait for a certain time), token usage is potentially associated with costs (see e.g. GPT-4). 

Table~\ref{tab:inference-time} provides an overview of the average inference time and the average tokens produced per model for all BPM tasks. 
Generally, higher inference times correlate with more tokens generated. For instance, Falcon2 and Yi-1.5 have longer inference times (2.18 and 1.74 seconds) and produce more tokens (49.89 and 32.02) compared to other models. However, more tokens do not necessarily mean better performance, as shown in previous experiments.
Llama3 has the fastest inference time at 0.19 seconds while generating 6.18 tokens on average, indicating a balance between speed and token generation. 
All models fall within an acceptable inference time range, but the trade-offs between speed, token count and performance highlight the importance of choosing models like Llama3 for optimal efficiency and effectiveness. Surprisingly, the smaller open-source models, such as Llama3-7b and Mixtral-8x7b, demonstrated competitive performance.
\begin{table}[ht]
\small
\centering
    \begin{tabular}{lcc}
    \hline
    \multirow{2}{*}{\textbf{LLM}} & \textbf{$\diameter$ inference} & \textbf{$\diameter$ completion} \\ 
     & \textbf{time (sec)} & \textbf{tokens} \\
    \hline
    \hline
    GPT-4&  1.06 & 5.14 \\ 
    \hline
    Phi3-14b & 0.31&9.39 \\
    Claude2-13b&0.40&10.57\\
    Falcon2-11b& 2.18&49.89\\
    Mixtral-8x7b&2.18&7.72\\
    Llama3-7b&\textbf{0.19}&6.18\\
    Yi-1.5-9b& 1.74&32.02\\
    \hline
    \end{tabular}
    \label{tab:inference-time}
\caption{Average inference time and average token production across different LLMs.}
\end{table}

\begin{figure*}[ht]
\centering
\includegraphics[width=0.65\textwidth, page=1]{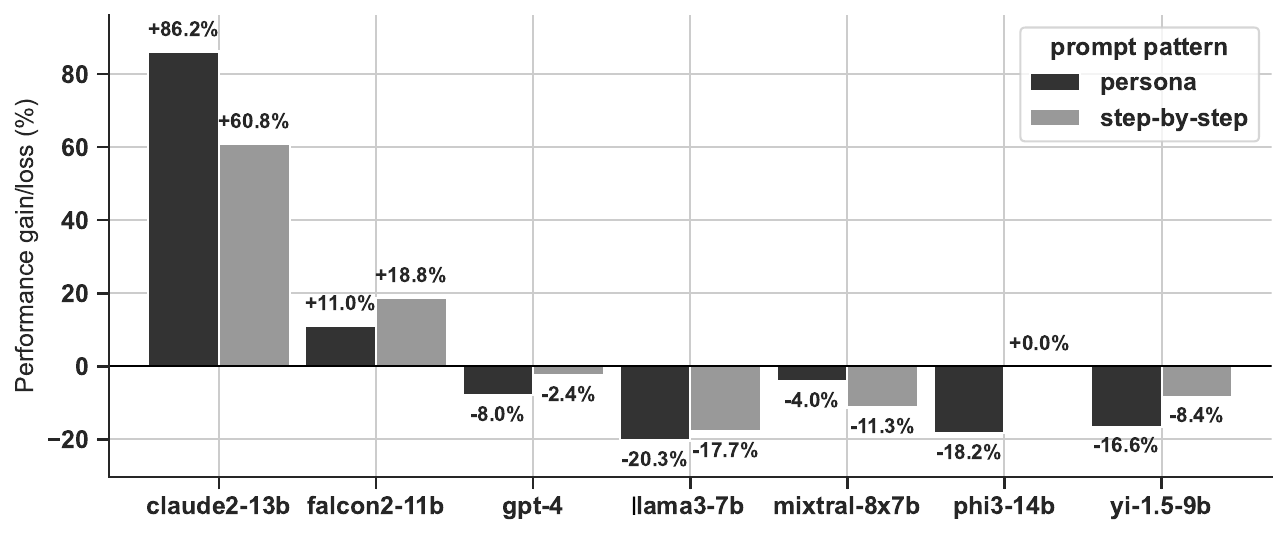}
\caption{Performance gain/loss by language models and prompt pattern (difference from persona baseline).}
\label{fig:prompt-templates}
\end{figure*}

\mypar{The importance of prompting}
Figure~\ref{fig:prompt-templates} shows the performance of the LLMs using different prompt templates across all BPM tasks. Specifically, it shows the percentage gain or loss when comparing persona prompt and step-by-step prompt templates to the few-shot example prompt template. 
Two main aspects can be observed. 
First, for models like phi3-14b, yi-1.5-9b, GPT-4, Mixtral-8x7b, and Llama3-7b, the the use of persona or step-by-step prompt templates leads to performance decreases. These models benefit from seeing multiple examples before generating outputs.
Second, models like falcon2-11b and Claude2-13b would benefit from using persona or step-by-step prompt templates, whereby the effect is stronger with the former.
This highlights that there is substantial variability in how LLMs respond to prompt templates.

\section{Implications}
\label{sec:implications}
The results reported in this paper have implications for both practice and research. 
In the following, we highlight these implications  by reviewing the three main insights from our paper.

    \mypar{GPT-4 is not best} While GPT-4 exhibited a stable performance across all BPM tasks, it is not the best model in terms of performance. Taking into account its relatively large inference time and the cost per generated token, we can conclude that it is not necessary for organization to invest into GPT-4 for the purpose of addressing BPM tasks. 
    
    \mypar{Model selection is important} Our results show that there are significant differences in performance between the models analyzed. Some models, as for instance Claude2, perform extraordinary well on one task but perform poorly on others. It is, hence, important for organizations to reflect on what BPM tasks need to be supported and select a suitable model respectively. One model that turned out to be stable across all tasks (besides GPT-4) is Llama3. 
    
    \mypar{Model size does not explain performance alone} There is a general consensus that performance of LLMs increases with model size \parencite{brown2020language, kaplan2020scaling}. However, our experiments reveal that model size alone is insufficient to explain performance differences. GPT-4 has over 1,760 billion parameters while the tested open-source models have between 8 and 46.7 billion parameters. While GPT-4 performs generally well across all tasks, also other, much smaller models yield comparable results.

Note that these insights cannot be easily generalized to the use of LLMs in other contexts. They do, however, highlight the necessity to properly reflect on the specific use cases and better understand when certain LLMs perform well. 

\section{Conclusion} 
\label{sec:conclusion}
In this paper, we investigated to what extent the results of existing LLM performance benchmarks can be transferred to the BPM domain. Specifically, we systematically compared the performance of open-source and close-source LLMs on four established BPM tasks.
First, despite showing a stable performance across all BPM tasks, GPT-4 is not the best performing model. Given its relatively large inference time and associated cost per generated token, we do not believe that the use of GPT-4 can be justified in the BPM context. Second, model selection is an important factor as the models exhibit varying strengths with respect to the different BPM tasks. Hence, organizations should properly reflect on their needs and use cases before deploying a particular LLM. Third, the size of an LLM cannot alone explain the performance. We showed that small models can perform equally well or even better than a large model, such as GPT-4. 

Naturally, our study is subject to a number of limitations. Most importantly, we cannot generalize the results reported in this paper to other domains or LLMs in general. We, however, believe that our results clearly show that the application of LLMs in specific domains requires a careful analysis and model selection. Furthermore, our results highlight the importance of domain-specific benchmarks and the need for further research in order to understand when and why a certain LLM should be selected.

\mypar{Acknowledgment} Part of this research was funded by the Deutsche Forschungsgemeinschaft (DFG, German Research Foundation) – Project No. 528177077.



\printbibliography

\end{document}